\documentclass[sigconf]{acmart}
\usepackage{booktabs} 
\usepackage{amsmath}
\usepackage{amsfonts}
\usepackage{comment}
\usepackage{footmisc}
\usepackage{mathrsfs}
\usepackage{amssymb}
\usepackage{graphicx}
\usepackage{subfigure}
\usepackage{multirow}
\usepackage{algorithm}
\usepackage{algorithmic}
\usepackage{multicol}  
\usepackage{tikz}
\usetikzlibrary{bayesnet}
\usepackage{enumitem}
\usepackage{array}
\usepackage{float}
\usepackage{hyperref}
\def\BibTeX{{\rm B\kern-.05em{\sc i\kern-.025em b}\kern-.08emT\kern-.1667em\lower.7ex\hbox{E}\kern-.125emX}}

\copyrightyear{2019}
\acmYear{2019}
\setcopyright{acmcopyright}
\acmConference[DRL4KDD '19]{The 1st Workshop on Deep Reinforcement Learning for Knowledge Discovery}{August 5, 2019}{Anchorage, AK, USA}

\acmBooktitle{The 1st Workshop on Deep Reinforcement Learning for Knowledge Discovery (DRL4KDD '19), August 5, 2019, Anchorage, AK, USA}

\begin{document}
\title{Deep Reinforcement Learning for List-wise Recommendations}
\author{Xiangyu Zhao}
\affiliation{
	\institution{Michigan State University}
}
\email{zhaoxi35@msu.edu}

\author{Liang Zhang}
\affiliation{
	\institution{JD.com}
}
\email{zhangliang16@jd.com}

\author{Long Xia}
\affiliation{
	\institution{JD.com}
}
\email{xialong@jd.com}

\author{Zhuoye Ding}
\affiliation{
	\institution{JD.com}
}
\email{dingzhuoye@jd.com}

\author{Dawei Yin}
\affiliation{
	\institution{JD.com}
}
\email{yindawei@acm.org}

\author{Jiliang Tang}
\affiliation{
	\institution{Michigan State University}
}
\email{tangjili@msu.edu}
\renewcommand{\shortauthors}{Xiangyu Zhao et al.}
\begin{abstract}
	Recommender systems play a crucial role in mitigating the problem of information overload by suggesting users' personalized items or services. The vast majority of traditional recommender systems consider the recommendation procedure as a static process and make recommendations following a fixed strategy. In this paper, we propose a novel recommender system with the capability of continuously improving its strategies during the interactions with users. We model the sequential interactions between users and a recommender system as a Markov Decision Process (MDP) and leverage Reinforcement Learning (RL) to automatically learn the optimal strategies via recommending trial-and-error items and receiving reinforcements of these items from users' feedbacks. In particular, we introduce an online user-agent interacting environment simulator, which can pre-train and evaluate model parameters offline before applying the model online. Moreover, we validate the importance of list-wise recommendations during the interactions between users and agent, and develop a novel approach to incorporate them into the proposed framework LIRD for list-wide recommendations. The experimental results based on a real-world e-commerce dataset demonstrate the effectiveness of the proposed framework. 
\end{abstract}


\maketitle

\section{Introduction}
\label{sec:introduction}
Recommender systems are intelligent E-commerce applications. They assist users in their information-seeking tasks by suggesting items (products, services, or information) that best fit their needs and preferences. Recommender systems have become increasingly popular in recent years, and have been utilized in a variety of domains including movies, music, books, point of interests, and social events\cite{zhao2019model,resnick1997recommender,ricci2011introduction,zhao2016exploring,zhao2018recommendations,zhao2018deep,zhao2018reinforcement,guo2016cosolorec}. Most existing recommender systems consider the recommendation procedure as a static process and make recommendations following a fixed greedy strategy. However, these approaches may fail given the dynamic nature of the users' preferences. Furthermore, the majority of existing recommender systems are designed to maximize the immediate (short-term) reward of recommendations, i.e., to make users order the recommended items, while completely overlooking whether these recommended items will lead to more likely or more profitable (long-term) rewards in the future~\cite{shani2005mdp}. 

In this paper, we consider the recommendation procedure as sequential interactions between users and recommender agent; and leverage Reinforcement Learning (RL) to automatically learn the optimal recommendation strategies. Recommender systems based on reinforcement learning have two advantages. First, they are able to continuously update their strategies during the interactions, until the system converges to the optimal strategy that generates recommendations best fitting users' dynamic preferences. Second, the optimal strategy is made by maximizing the expected long-term cumulative reward from users. Therefore, the system can identify the item with a small immediate reward but making big contribution to the rewards for future recommendations.

\begin{figure*}
	\centering
	\includegraphics[width=166mm]{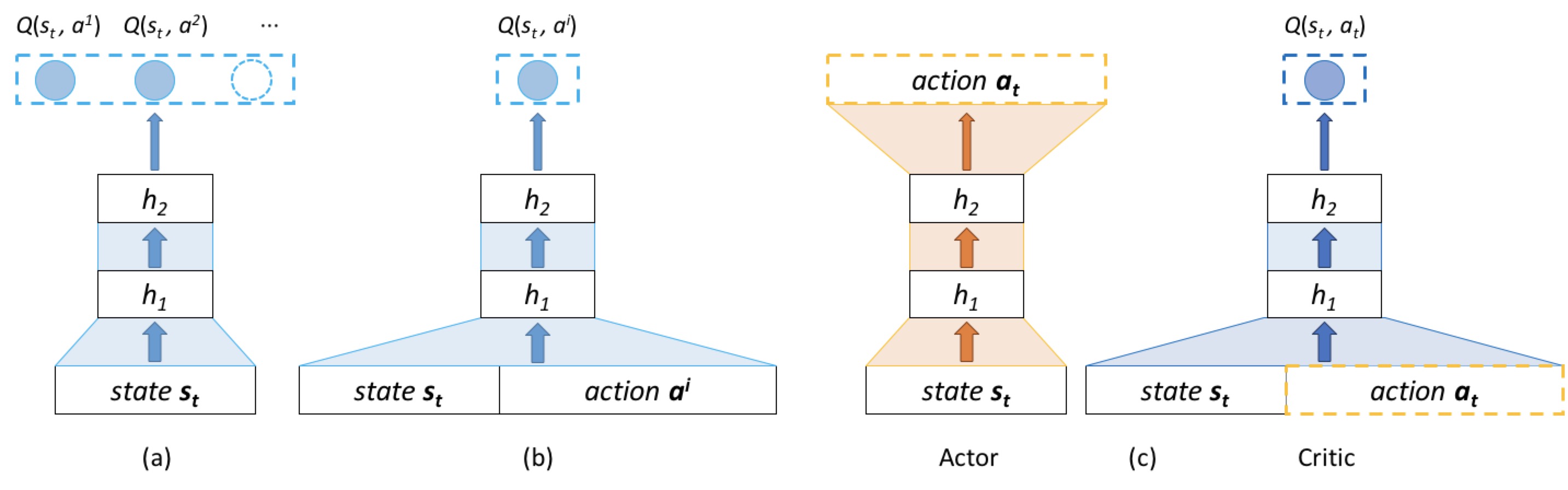}
	\caption{DQN architecture selection.}
	\label{fig:selection}
\end{figure*}

Efforts have been made on utilizing reinforcement learning for recommender systems, such as POMDP\cite{shani2005mdp} and Q-learning\cite{taghipour2008hybrid}. However, these methods may become inflexible with the increasing number of items for recommendations. This prevents them to be adopted by practical recommender systems. Thus, we leverage Deep Reinforcement Learning\cite{lillicrap2015continuous} with (adapted) artificial neural networks as the non-linear approximators to estimate the action-value function in RL. This model-free reinforcement learning method does not estimate the transition probability and not store the Q-value table. This makes it flexible to support huge amount of items in recommender systems. 

\subsection{List-wise Recommendations}
Users in practical recommender systems are typically recommended a list of items at one time.  List-wise recommendations are more desired in practice since they allow the systems to provide diverse and complementary options to their users.  For list-wise recommendations, we have a list-wise action space, where each action is a set of multiple interdependent sub-actions (items). Existing reinforcement learning recommender methods also could recommend a list of items. For example, DQN\cite{mnih2013playing} can calculate Q-values of all recalled items separately, and recommend a list of items with highest Q-values. However, these approaches recommend items based on one same state, and ignore relationship among the recommended items. As a consequence, the recommended items are similar. In practice, a bundling with complementary items may receive higher rewards than recommending all similar items. For instance, in real-time news feed recommendations, a user may want to read diverse topics of interest, and an action (i.e. recommendation) from the recommender agent would consist of a set of news articles that are not all similar in topics\cite{yue2011linear}. Therefore, in this paper, we propose a principled approach to capture relationship among recommended items and generate a list of complementary items to enhance the performance.

\subsection{Architecture Selection}
Generally, there exist two Deep Q-learning architectures, shown in Fig.\ref{fig:selection} (a)(b). Traditional deep Q-learning adopts the first architecture as shown in Fig.\ref{fig:selection}(a), which inputs only the state space and outputs Q-values of all actions. This architecture is suitable for the scenario with high state space and small action space, like playing Atari\cite{mnih2013playing}. However, one drawback is that it cannot handle large and dynamic action space scenario, like recommender systems. The second Q-learning architecture, shown Fig.\ref{fig:selection}(b), treats the state and the action as the input of Neural Networks and outputs the Q-value corresponding to this action. This architecture does not need to store each Q-value in memory and thus can deal with large action space or even continuous action space. A challenging problem of leveraging the second architecture is temporal complexity, i.e., this architecture computes Q-value for all potential actions, separately. To tackle this problem, in this paper, our recommending policy builds upon the Actor-Critic framework\cite{sutton1998reinforcement}, shown in Fig.\ref{fig:selection} (c). The Actor inputs the current state and aims to output the parameters of a state-specific scoring function. Then the RA scores all items and selects an item with the highest score. Next, the Critic uses an approximation architecture to learn a value function (Q-value), which is a judgment of whether the selected action matches the current state. Note that Critic shares the same architecture with the DQN in Fig.\ref{fig:selection}(b).  Finally, according to the judgment from Critic, the Actor updates its' policy parameters in a direction of recommending performance improvement to output properer actions in the following iterations. This architecture is suitable for large action space, while can also reduce redundant computation simultaneously.

\subsection{Online Environment Simulator}
Unlike the Deep Q-learning method applied in playing Online Game like Atari, which can take arbitrary action and obtain timely feedback/reward, the online reward is hard to obtain before the recommender system is applied online. In practice, it is necessary to pre-train parameters offline and evaluate the model before applying it online, thus how to train our framework and evaluate the performance of our framework offline is a challenging task. To tackle this challenge, we propose an online environment simulator, which inputs current state and a selected action and outputs a simulated online reward, which enables the framework to train the parameters offline based on the simulated reward. More specifically, we build the simulator by users' historical records. The intuition is no matter what algorithms a recommender system adopt, given the same state ( or a user's historical records) and the same action (recommending the same items to the user), the user will make the same feedbacks to the items.

To evaluate the performance of a recommender system before applying it online, a practical way is to test it based on users' historical clicking/ordering records. However, we only have the ground truth feedbacks (rewards) of the existing items in the users' historical records, which are sparse compared with the enormous item space of current recommender system. Thus we cannot get the feedbacks (rewards) of items that are not in users' historical records. This may result in inconsistent results between offline and online measurements. Our proposed online environment simulator can also mitigate this challenge by producing simulated online rewards given any state-action pair, so that the recommender system can rate items from the whole item space. Based on offline training and evaluation, the well trained parameters can be utilized as the initial parameters when we launch our framework online, which can be updated and improved via on-policy exploitation and exploration.

\subsection{Our Contributions}
We summarize our major contributions as follows: 
\begin{itemize}[leftmargin=*]
	\item We build an online user-agent interacting environment simulator, which is suitable for offline parameters pre-training and evaluation before applying a recommender system online;
	\item We propose a {\bf LI}st-wise {\bf R}ecommendation framework based on {\bf D}eep reinforcement learning LIRD, which can be applied in scenarios with large and dynamic item space and can reduce redundant computation significantly; and 
	\item We demonstrate the effectiveness of the proposed framework in a real-world e-commerce dataset and validate the importance of list-wise recommendation for accurate recommendations.
\end{itemize}

The rest of this paper is organized as follows. In Section 2, we first formally define the problem of recommender system via reinforcement learning. Then, we provide approaches to model the recommending procedure as a sequential user-agent interactions and introduce details about employing Actor-Critic framework to automatically learn the optimal recommendation strategies via a online simulator. Section 3 carries out experiments based on real-word e-commerce site and presents experimental results.  Section 4 briefly reviews related work. Finally, Section 5 concludes this paper and discusses our future work.

\section{The Proposed Framework}
\label{sec:framework}

In this section, we first formally define notations and the problem of recommender system via reinforcement learning. Then we build an online user-agent interaction environment simulator. Next, we propose an Actor-Critic based reinforcement learning framework under this setting. Finally, we discuss how to train the framework via users' behavior log and how to utilize the framework for list-wise recommendations. 

\subsection{Problem Statement}
\label{sec:problem} 

We study the recommendation task in which a recommender agent (RA) interacts with environment $\mathcal{E}$ (or users) by sequentially choosing recommendation items over a sequence of time steps, so as to maximize its cumulative reward. We model this problem as a Markov Decision Process (MDP), which includes a sequence of states, actions and rewards. More formally, MDP consists of a tuple of five elements $(\mathcal{S}, \mathcal{A}, \mathcal{P}, \mathcal{R}, \gamma)$ as follows:

\begin{itemize}[leftmargin=*]
	\item {\bf State space $\mathcal{S}$}: A state $s_t = \{s_t^{1}, \cdots, s_t^{N}\} \in \mathcal{S}$ is defined as the browsing history of a user, i.e., previous $N$ items that a user browsed before time $t$. The items in $s_t$ are sorted in chronological order.  
	\item {\bf Action space $\mathcal{A}$}:  An action $a_t = \{a_t^1,  \cdots, a_t^K\}\in \mathcal{A}$ is to recommend a list of items to a user at time $t$ based on current state $s_t$, where $K$ is the number of items the RA recommends to user each time. 
	\item {\bf Reward $\mathcal{R}$}: After the recommender agent takes an action $a_t$ at the state $s_t$, i.e., recommending a list of items to a user, the user browses these items and provides her feedback. She can skip (not click), click, or order these items, and the agent receives immediate reward $r(s_t,a_t)$ according to the user's feedback.
	\item {\bf Transition probability $\mathcal{P}$}: Transition probability $p(s_{t+1}|s_t,a_t)$ defines the probability of state transition from $s_t$ to $s_{t+1}$ when RA takes action $a_t$. We assume that the MDP satisfies $p(s_{t+1}|s_t,a_t,...,$ $s_1,a_1) = p(s_{t+1}|s_t,a_t)$. If user skips all the recommended items, then the next state $s_{t+1} = s_{t}$; while if the user clicks/orders part of items, then the next state $s_{t+1}$ updates.  More details will be shown in following subsections.
	\item {\bf Discount factor $\gamma$}: $\gamma \in [0,1]$ defines the discount factor when we measure the present value of future reward. In particular, when $\gamma = 0$, RA only considers the immediate reward. In other words, when $\gamma = 1$, all future rewards can be counted fully into that of the current action.
\end{itemize}

In practice, only using discrete indexes to denote items is not sufficient since we cannot know the relations between different items only from indexes. One common way is to use extra information to represent items. For instance, we can use the attribute information like brand, price, sale per month, etc. Instead of extra item information, in this paper, we use the user-agent interaction information, i.e., users' browsing history. We treat each item as a word and the clicked items in one recommendation session as a sentence. Then, we can obtain dense and low-dimensional vector representations for items via word embedding\cite{levy2014neural}.

\begin{figure}[H]
	\centering
	\includegraphics[width=81mm]{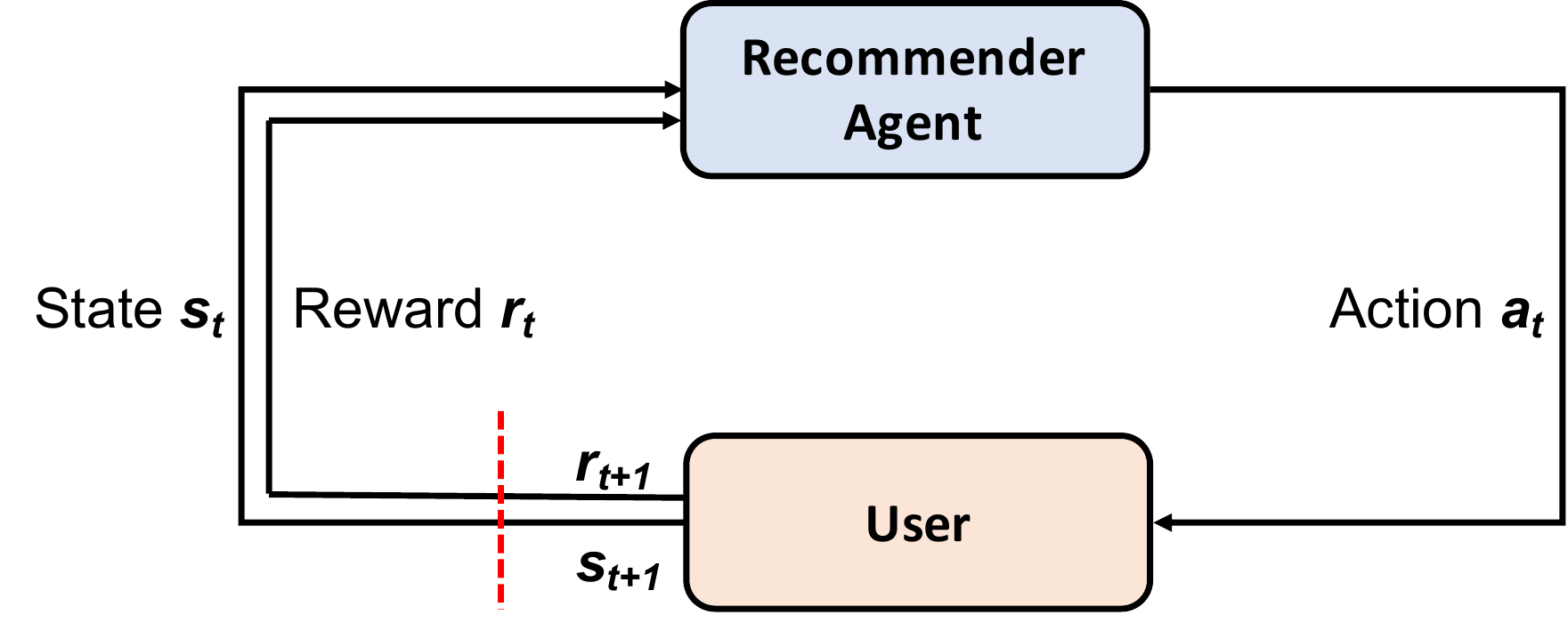}
	\caption{The agent-user interactions in MDP.}
	\label{fig:interaction}
\end{figure}

Figure \ref{fig:interaction} illustrates the agent-user interactions in MDP. By interacting with the environment (users), recommender agent takes actions (recommends items) to users in such a way that maximizes the expected return, which includes the delayed rewards. We follow the standard assumption that delayed rewards are discounted by a factor of $\gamma$ per time-step.

With the notations and definitions above, the problem of list-wise item recommendation can be formally defined as follows: \textit{Given the historical MDP, i.e., $(\mathcal{S}, \mathcal{A}, \mathcal{P}, \mathcal{R}, \gamma)$, the goal is to find a recommendation policy $\pi:\mathcal{S} \rightarrow \mathcal{A}$, which can  maximize the cumulative reward for the recommender system.}

\subsection{Online User-Agent Interaction Environment Simulator}
\label{sec:simulator}
To tackle the challenge of training our framework and evaluating the performance of our framework offline, in this subsection, we propose an online user-agent interaction environment simulator. In the online recommendation procedure, given the current state $s_t$, the RA recommends a list of items $a_t$ to a user, and the user browses these items and provides her feedbacks, i.e., skip/click/order part of the recommended items. The RA receives immediate reward $r(s_t,a_t)$ according to the user's feedback. To simulate the aforementioned online interaction procedures, the task of simulator is to predict a reward based on current state and a selected action, i.e., $f: (s_t,a_t) \rightarrow r_t$. 

According to collaborative filtering techniques, users with similar interests will make similar decisions on the same item. With this intuition, we match the current state and action to existing historical state-action pairs, and stochastically generate a simulated reward. To be more specific, we first build a memory $\mathcal{M} = \{m_1, m_2, \cdots\}$ to store users' historical browsing history, where $m_i$ is a user-agent interaction triple $((s_i,a_i) \rightarrow r_i)$. The procedure to build the online simulator memory is illustrated in Algorithm \ref{alg:simulator}. Given a historical recommendation session $\{a_1, \cdots, a_L\}$, we can observe the initial state $s_0 = \{s_0^1, \cdots, s_0^N\}$ from the previous sessions (line 2). Each time we observe $K$ items in temporal order (line 3), where ``$l =1, L; K$" means that each iteration we will move forward a window of $K$. We can observe the current state (line 4), current $K$ items (line 5), and the user's feedbacks for these items (line 6). Then we store triple $((s,a)\rightarrow r)$ in memory(line-7). Finally we update the state (lines 8-13), and move to the next $K$ items. Since we keep a fixed length state $s = \{s^1, \cdots, s^N\}$, each time a user clicked/ordered some items in the recommended list, we add these items to the end of state and remove the same number of items in the top of the state. For example, the RA recommends a list of five items $\{a_{1}, \cdots, a_{5}\} $ to a user, if the user clicks $a_{1}$ and orders $a_{5}$, then update $s = \{s^{3}, \cdots, s^{N}, a_{1}, a_{5}\}$.

\begin{algorithm}
	\caption{\label{alg:simulator} Building Online Simulator Memory.}
	\raggedright
	{\bf Input}: Users' historical sessions $B$, and the length of recommendation list $K$.\\
	{\bf Output}:Simulator Memory $\mathcal{M}$\\
	\begin{algorithmic} [1]
		
		\FOR{$session =1, B$}
			\STATE Observe initial state $s_0 = \{s_0^1, \cdots, s_0^N\}$
			\FOR{$item \, order\, l =1, L; K$}
			
				\STATE Observe current state $s = \{s^1, \cdots, s^N\}$
				\STATE Observe current action list $a = \{a_{l}, \cdots, a_{l+K-1}\}$
				\STATE Observe current reward list  $r = \{r_{l}, \cdots, r_{l+K-1}\}$
				\STATE Add the triple $((s,a)\rightarrow r)$ in $\mathcal{M}$

				\FOR{$k =0, K-1$}
					\IF{$r_{l+k} > 0$}	
						\STATE Remove the first item in $s$
						\STATE Add the item $a_{l+k}$ in the bottom of $s$
					\ENDIF
				\ENDFOR
			\ENDFOR
		\ENDFOR
		
		\RETURN{$\mathcal{M}$}
	\end{algorithmic}
\end{algorithm}

Then we calculated the similarity of the current state-action pair, say $p_t (s_t, a_t)$, to each existing historical state-action pair in the memory. In this work, we adopt cosine similarity as:
\begin{equation}\label{equ:cosine}
Cosine(p_t, m_i)=\alpha\frac{s_t s_i^\top}{ \| s_t \| \| s_i \|} + (1-\alpha)\frac{a_t a_i^\top}{ \| a_t \| \| a_i \|} ,
\end{equation}
where the first term measures the state similarity and the second term evaluates the action similarity. Parameter $\alpha$ controls the balance of two similarities. Intuitively, with the increase of similarity between $p_t$ and $m_i$, there is a higher chance $p_t$ mapping to the reward $r_i$. Thus the probability of $p_t \rightarrow r_i$ can be defined as follows:

\begin{equation}\label{equ:probability}
P(p_t \rightarrow r_i)=\frac{Cosine(p_t, m_i)}{\sum_{m_j \in \mathcal{M}}Cosine(p_t, m_j)}, 
\end{equation}
then we can map the current state-action pair $p_t$ to a reward according the above probability. The major challenge of this projection is the computation complexity, i.e., we must compute pair-wise similarity between $p_t$ and each $m_i \in \mathcal{M}$. To tackle this challenge, we first group users' historical browsing history according to the rewards. Note that the number of reward permutation is typically limited. For example, the RA recommends two items to user each time, and the reward of user skip/click/order an item is 0/1/5, then the permutation of two items' rewards is 9, i.e., $\mathcal{U}  = \{\mathcal{U}_1, \cdots, \mathcal{U}_9\}= \{(0,0), (0,1), (0,5), (1,0), (1,1), (1,5), (5,0), (5,1), (5,5)\}$, which is much smaller than the total number of historical records. Then probability of mapping $p_t$ to $\mathcal{U}_x$ can be computed as follows:
\begin{small}
\begin{equation}\label{equ:probability1}
\begin{aligned}
P(p_t \rightarrow \mathcal{U}_x) & =\frac{\sum_{r_i = \mathcal{U}_x}Cosine(p_t, m_i)}{\sum_{m_j \in \mathcal{M}}Cosine(p_t, m_j)}\\
&= \frac{\alpha \frac{s_t}{ \| s_t \|}\cdot \sum\limits_{r_i  = \mathcal{U}_x}\frac{s_i^\top}{\| s_i \|} + (1-\alpha) \frac{a_t}{ \| a_t \|}\cdot \sum\limits_{r_i  = \mathcal{U}_x}\frac{a_i^\top}{\| a_i \|}} {\sum\limits_{\mathcal{U}_y \in \mathcal{U}}\bigg(\alpha \frac{s_t}{ \| s_t \|}\cdot \sum\limits_{r_j  = \mathcal{U}_y}\frac{s_j^\top}{\| s_j \|} + (1-\alpha) \frac{a_t}{ \| a_t \|}\cdot \sum\limits_{r_j  = \mathcal{U}_y}\frac{a_j^\top}{\| a_j \|}\bigg)}\\
&=\frac{\mathcal{N}_x\cdot\bigg(\alpha \frac{s_t\bar{s_x}^\top}{ \| s_t \| } + (1-\alpha) \frac{a_t\bar{a_x}^\top}{ \| a_t \|}\bigg)} {\sum\limits_{\mathcal{U}_y \in \mathcal{U}}\mathcal{N}_y \cdot\bigg(\alpha \frac{s_t\bar{s_y}^\top}{ \| s_t \|} + (1-\alpha) \frac{a_t\bar{a_y}^\top }{ \| a_t \|}\bigg)}
\end{aligned}
\end{equation}
\end{small}
 where we assume that $r_i$ is a reward list containing user's feedbacks of the recommended items, for instance $r_i = (1,5)$. $\mathcal{N}_x$ is the size of users' historical browsing history group that $r  = \mathcal{U}_x$. $\bar{s_x} $ and $\bar{a_x}$ are the average state vector and average action vector for $r  = \mathcal{U}_x$, i.e., $\bar{s_x} = \frac{1}{\mathcal{N}_x}\sum_{r_i  = \mathcal{U}_x}s_i/\|s_i\|$, and $\bar{a_x} = \frac{1}{\mathcal{N}_x}\sum_{r_i  = \mathcal{U}_x}a_i/\|a_i\|$.  The simulator only needs to pre-compute the $\mathcal{N}_x$, $\bar{s_x} $ and $\bar{a_x}$, and can map $p_t$ to a reward list $\mathcal{U}_x$ according to the probability in Eq.(\ref{equ:probability1}). In practice, RA updates $\mathcal{N}_x$, $\bar{s_x} $ and $\bar{a_x}$ every 1000 episodes. As $|\mathcal{U}|$ is much smaller than the total number of historical records, Eq.(\ref{equ:probability1}) can map $p_t$ to a reward list $\mathcal{U}_x$ efficiently.
  
In practice, the reward is usually a number, rather than a vector. Thus if the $p_t$ is mapped to $\mathcal{U}_x$, we calculate the overall reward $r_t$ of the whole recommended list as follows:
\begin{equation}\label{equ:rewrad}
r_t=\sum_{k= 1}^K \Gamma^{k-1}\mathcal{U}_x^k, 
\end{equation}
where $k$ is the order that an item in the recommended list and $K$ is the length of the recommended list, and $\Gamma \in (0, 1]$. The intuition of Eq.(\ref{equ:rewrad}) is that reward in the top of recommended list has a higher contribution to the overall rewards, which force RA arranging items that user may order in the top of the recommended list.

\begin{figure*}[t]
	\centering
	\includegraphics[width=136mm]{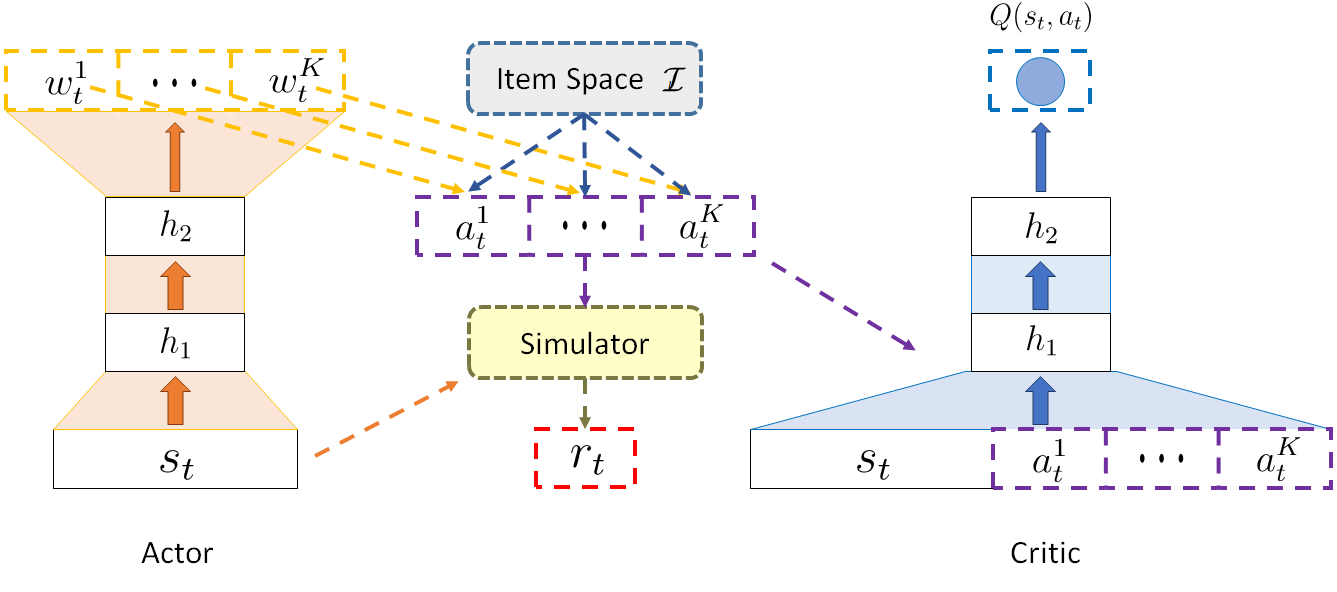}
	\caption{An illustration of the proposed framework with online simulator.\label{fig:framework}}
\end{figure*}

\subsection{The Actor Framework}
\label{sec:actor}
In this subsection, we propose the list-wise item recommending procedure, which consists of two steps, i.e., 1) state-specific scoring function parameter generating, and 2) action generating. Current practical recommender systems rely on a scoring or rating system which is averaged across all users ignoring specific demands of a user. These approaches perform poorly in tasks where there is large variation in users' interests. To tackle this problem, we present a state-specific scoring function, which rates items according to user's current state. 

In the previous section, we have defined the state $s$ as the whole browsing history, which can be infinite and inefficient. A better way is to only consider the positive items, e.g., previous 10 clicked/ ordered items. A good recommender system should recommend the items that users prefer the most. The positive items represent key information about users' preferences, i.e., which items the users prefer to. Thus, we only consider them for state-specific scoring function. 

Our state-specific scoring function parameter generating step maps the current state $s_t = \{s_t^{1}, \cdots, s_t^{N}\}$ to a list of weight vectors $\mathbf{w}_t = \{\mathbf{w}_t^{1}, \cdots, \mathbf{w}_t^{K}\}$ as  follows:
\begin{equation}\label{equ:scoring}
f_{\theta^\pi}:  s_t \rightarrow \mathbf{w}_t
\end{equation}
where $f_{\theta^\pi}$ is a function parametrized by $\theta^\pi$, mapping from the state space to the weight representation space. Here we choose deep neural networks as the parameter generating function. 

Next we present the action-generating step based on the aforementioned scoring function parameters. Without the loss of generality, we assume that the scoring function parameter $\mathbf{w}_t^k$ and the embedding $\mathbf{e}_i$ of $i^{th}$ item from the item space $\mathcal{I}$ is linear-related as: 
\begin{equation}\label{equ:scores}
socre_i = \mathbf{w}_t^k \mathbf{e}_i^\top.
\end{equation}

Note that it is straightforward to extend it with non-linear relations. Then after computing scores of all items, the RA selects an item with highest score as the sub-action $a_t^k$ of action $a_t$.  We present list-wise item recommendation algorithm in Algorithm \ref{alg:actor}.  

\begin{algorithm}
	\caption{\label{alg:actor} List-Wise Item Recommendation Algorithm.}
	\raggedright
	{\bf Input}: Current state $s_t$, Item space $\mathcal{I}$, the length of recommendation list $K$.\\
	{\bf Output}:Recommendation list  $a_t$.\\
	\begin{algorithmic} [1]
		\STATE Generate $\mathbf{w}_t = \{\mathbf{w}_t^{1}, \cdots, \mathbf{w}_t^{K}\}$ according Eq.(\ref{equ:scoring})
		\STATE \textbf{for} $k =1, K$ \textbf{do}
		\STATE \quad Score items in $\mathcal{I}$ according Eq.(\ref{equ:scores})
		\STATE \quad Select the an item with highest score as $a_t^k$
		\STATE \quad Add item $a_t^k$ in the bottom of $a_t$
		\STATE \quad Remove item $a_t^k$ from $\mathcal{I}$
		\STATE \textbf{end for}
		\STATE \textbf{return} $a_t$
	\end{algorithmic}
\end{algorithm}

The Actor first generates a list of weight vectors (line 1). For each weight vector, the RA scores all items in the item space (line 3), selects the item with highest score (line 4), and then adds this item at the end of the recommendation list. Finally the RA removes this item from the item space, which prevents recommending the same item to the recommendation list.

\subsection{The Critic Framework}
\label{sec:traditional}
The Critic is designed to leverage an approximator to learn an action-value function $Q(s_t, a_t)$, which is a judgment of whether the action $a_t$ generated by Actor matches the current state $s_t$.  Then, according $Q(s_t, a_t)$, the Actor updates its' parameters in a direction of improving performance to generate proper actions in the following iterations.  Many applications in reinforcement learning make use of the optimal action-value function $Q^*(s_t, a_t)$. It is the maximum expected return achievable by the optimal policy, and should follow the Bellman equation \cite{bellman2013dynamic} as:

\begin{equation}\label{equ:Q*sa}
	Q^{*}(s_t, a_t)=\mathbb{E}_{s_{t+1}} \, \big[r_t+\gamma\max_{a_{t+1}}Q^{*}(s_{t+1}, a_{t+1})|s_t, a_t\big].
\end{equation}

In practice, to select an optimal $a_{t+1}$, $|\mathcal{A}|$ evaluations are necessary for the inner operation $\max$. This prevents Eq.(\ref{equ:Q*sa}) to be adopted in practical recommender systems with the enormous action space. However, the Actor architectures proposed in Section \ref{sec:actor} outputs a deterministic action for Critic, which avoids the aforementioned computational cost of $|\mathcal{A}|$ evaluations in Eq.(\ref{equ:Q*sa}) as follows:
\begin{equation}\label{equ:Qsa}
Q(s_t, a_t)=\mathbb{E}_{s_{t+1}} \, \big[r_t+\gamma Q(s_{t+1}, a_{t+1})|s_t, a_t\big].
\end{equation}
where the Q-value function $Q(s_t, a_t)$ is the expected return based on state $s_t$ and the action $a_t$.

In real recommender systems, the state and action spaces are enormous, thus estimating the action-value function $Q(s, a)$ for each state-action pair is infeasible. In addition, many state-action pairs may not appear in the real trace such that it is hard to update their values. Therefore, it is more flexible and practical to use an approximator function to estimate the action-value function, i.e., $Q(s,a) \approx Q(s, a; \theta^\mu)$ . In practice, the action-value function is usually highly nonlinear. Deep neural networks are known as excellent approximators for non-linear functions. In this paper, We refer to a neural network function approximator with parameters $\theta^\mu$ as deep $\mathrm{Q}$-network (DQN). A DQN can be trained by minimizing a sequence of loss functions $L(\theta^\mu)$ as
\begin{equation}\label{equ:L}
	L(\theta^\mu)=\mathbb{E}_{s_t, a_t,r_t,s_{t+1}}\big[(y_t-Q(s_t, a_t;\theta^\mu))^{2}\big],
\end{equation}
where $y_t= \mathbb{E}_{s_{t+1}}[r_t+\gamma Q'(s_{t+1},\ a_{t+1};\theta^{\mu'})|s_{t}, a_{t}]$ is the target for the current iteration. The parameters from the previous iteration $\theta^{\mu'}$ are fixed when optimizing the loss function $L(\theta^\mu)$. In practice, it is often computationally efficient to optimize the loss function by stochastic gradient descent, rather than computing the full expectations in the above gradient.


\subsection{The Training Procedure}
\label{sec:training}
An illustration of the proposed user-agent online interaction simulator and deep reinforcement recommending LIRD framework is demonstrated in Figure~\ref{fig:framework}. Next, we discuss the parameters training procedures.  In this work, we utilize DDPG algorithm\cite{lillicrap2015continuous} to train the parameters of the proposed framework. The training algorithm for the proposed framework DEV is presented in Algorithm \ref{alg:training}.
\begin{small}
	\begin{algorithm}
		\caption{\label{alg:training} Parameters Training for DEV with DDPG.}
		\raggedright
		\begin{algorithmic} [1]
			\STATE Initialize actor network $f_{\theta^\pi}$ and critic network $Q(s,a|\theta^{\mu})$ with random weights	
			\STATE Initialize target network $f'$ and $Q'$ with weights $\theta^{\pi'}\leftarrow\theta^{\pi}, \theta^{\mu'}\leftarrow\theta^{\mu}$
			
			\STATE Initialize the capacity of replay memory $\mathcal{D}$
			\FOR{$session =1, M$}
			\STATE Reset the item space $\mathcal{I}$
			\STATE  Initialize state $s_{0}$ from previous sessions
			\FOR{$t=1, T$}
			\STATE \textbf{Stage 1: Transition Generating Stage}
			\STATE  Select an action $a_t = \{a_t^1,  \cdots, a_t^K\}$ according \textbf{Alg.\ref{alg:actor}}
			\STATE  Execute action $a_t$ and observe the reward list $\{r_t^1,  \cdots, r_t^K\}$ for each item in $a_t$
			\STATE Set $s_{t+1} = s_{t}$
			\FOR{$k=1, K$} 
			\IF{$r_{t}^k > 0$}
			\STATE  Add $a_{t}^k$ to the end of $s_{t+1}$
			\STATE  Remove the first item of $s_{t+1}$
			\ENDIF
			\ENDFOR
			\STATE Compute overall reward $r_t$ according Eq. (\ref{equ:rewrad})
			\STATE  Store transition $(s_{t}, a_{t},  r_{t},s_{t+1})$ in $\mathcal{D}$
			\STATE Set $s_{t} = s_{t+1}$
			\STATE \textbf{Stage 2: Parameter Updating Stage}
			\STATE  Sample minibatch of $\mathcal{N}$ transitions $(s, a, r, s')$ from $\mathcal{D}$
			\STATE Generate $a'$  by target Actor network according \textbf{Alg.\ref{alg:actor}}
			\STATE  Set $y=  r+\gamma Q'(s',a';\theta^{\mu'})  $
			\STATE  Update Critic by minimizing $\big(y-Q(s, a;\theta^\mu)\big)^{2}$ according to:
			$$\nabla_{\theta^\mu}L(\theta^\mu) \approx\frac{1}{\mathcal{N}}\big[(y-Q(s, a;\theta^\mu))\nabla_{\theta^\mu}Q(s, a;\theta^\mu)\big]$$
			
			\STATE  Update the Actor using the sampled policy gradient: $$\nabla_{\theta^{\pi}}f_{\theta^\pi}\approx\frac{1}{\mathcal{N}}\sum_{i}\nabla_{a}Q(s, a|\theta^{\mu})\nabla_{\theta^{\pi}}f_{\theta^\pi}(s)$$
			\STATE Update the Critic target networks: $$\theta^{\mu'}\leftarrow\tau\theta^{\mu}+(1-\tau)\theta^{\mu'}$$
			\STATE Update the Actor target networks:
			$$
			\theta^{\pi'}\leftarrow\tau\theta^{\pi}+(1-\tau)\theta^{\pi'}
			$$
			\ENDFOR 
			\ENDFOR 
		\end{algorithmic}
	\end{algorithm}
\end{small}

In each iteration, there are two stages, i.e., 1) transition generating stage (lines 8-20), and 2) parameter updating stage (lines 21-28). For transition generating stage (line 8): given the current state $s_t$, the RA first recommends a list of items $a_t = \{a_t^1,  \cdots, a_t^K\}$ according to Algorithm~\ref{alg:actor} (line 9); then the agent observes the reward $r_t$ from simulator (line 10) and updates the state to $s_{t+1}$ (lines 11-17) following the same strategy in Algorithm \ref{alg:simulator}; and finally the recommender agent stores transitions $(s_t,a_t,r_t,s_{t+1})$ into the memory $\mathcal{D}$ (line 19), and set $s_{t} = s_{t+1}$(line 20). For parameter updating stage: the recommender agent samples mini-batch of transitions $(s, a, r, s')$ from $\mathcal{D}$ (line 22), and then updates parameters of Actor and Critic (lines 23-28) following a standard DDPG procedure~\cite{lillicrap2015continuous}.

In the algorithm, we introduce widely used techniques to train our framework. For example, we utilize a technique known as {\it experience replay} \cite{lin1993reinforcement} (lines 3,22), and introduce separated evaluation and target networks~\cite{mnih2013playing}(lines 2,23), which can help smooth the learning and avoid the divergence of parameters. For the soft target updates of target networks(lines 27,28), we used $\tau = 0.001$. Moreover, we leverage {\it prioritized sampling strategy}~\cite{moore1993prioritized} to assist the framework learning from the most important historical transitions.

\subsection{The Testing Procedure}

\label{sec:testing}

After framework training stage, RA gets well-trained parameters, say $\Theta^{\pi}$ and $\Theta^{\mu}$. Then we can do framework testing on simulator environment. The model testing also follows Algorithm \ref{alg:training}, i.e., the parameters continuously updates during the testing stage, while the major difference from training stage is before each recommendation session, we reset the parameters back to $\Theta^{\pi}$ and $\Theta^{\mu}$, for the sake of fair comparison between each session. We can artificially control the length of recommendation session to study the short-term and long-term performance.
\section{Experiments}
\label{sec:experiments}
In this section, we conduct extensive experiments with a dataset from a real e-commerce site to evaluate the effectiveness of the proposed framework. We mainly focus on two questions: (1) how the proposed framework performs compared to representative baselines; and (2) how the list-wise strategy contributes to the performance. We first introduce experimental settings. Then we seek answers to the above two questions. Finally, we study the impact of important parameters on the performance of the proposed framework. 

\subsection{Experimental Settings}
\label{sec:experimental_settings}

We evaluate our method on a dataset of July, 2017 from a real e-commerce site. We randomly collect 100,000 recommendation sessions (1,156,675 items) in temporal order, and use the first 70\% sessions as the training set and the later 30\% sessions as the testing set. For a given session, the initial state is collected from the previous sessions of the user. In this paper, we leverage $N = 10$ previously clicked/ordered items as the positive state. Each time the RA recommends a list of $K = 4$ items to users. The reward $r$ of skipped/clicked/ordered items are empirically set as 0, 1, and 5, respectively. The dimension of the item embedding is 50, and we set the discounted factor $\gamma = 0.75$. For the parameters of the proposed framework such as $K$ and $\gamma$, we select them via cross-validation. Correspondingly, we also do parameter-tuning for baselines for a fair comparison. We will discuss more details about parameter selection for the proposed framework in the following subsections. 

To evaluate the performance of the proposed framework, we select \textbf{MAP}~\cite{turpin2006user} and \textbf{NDCG}~\cite{jarvelin2002cumulated} as the metrics to measure the performance. The difference of ours from traditional Learn-to-Rank methods is that we rank both clicked and ordered items together, and set them by different rewards, rather than only rank clicked items as that in Learn-to-Rank problems.

\subsection{Performance Comparison for Item Recommendations}
\label{sec:ev_overall}

To answer the the first question, we compare the proposed framework with the following representative baseline methods: 

\begin{itemize}[leftmargin=*]
	\item \textbf{CF}: Collaborative filtering\cite{breese1998empirical} is a method of making automatic predictions about the interests of a user by collecting preference information from many users, which is based on the hypothesis that people often get the best recommendations from someone with similar tastes to themselves. 
	\item \textbf{FM}: Factorization Machines\cite{rendle2010factorization} combine the advantages of support vector machines with factorization models. Compared with matrix factorization, higher order interactions can be modeled using the dimensionality parameter.
	\item \textbf{DNN}: We choose a deep neural network with back propagation technique as a baseline to recommend the items in a given session. The input of DNN is the embeddings of users' historical clicked/ordered items. We train the DNN to output the next recommended item.
	\item \textbf{RNN}: This baseline utilizes the basic RNN to predict what user will buy next based on the clicking/ordering histories. To minimize the computation costs, it only keeps a  finite number of the latest states.
    \item \textbf{DQN}: We use a Deep Q-network\cite{mnih2013playing} with embeddings of users' historical clicked/ordered items (state) and a recommended item (action) as input, and train this baseline following Eq. \ref{equ:Q*sa}. Note that the DQN shares the same architecture with the Critic in our framework. 
\end{itemize}

\begin{figure}[t]
	\centering
	\includegraphics[width=81mm]{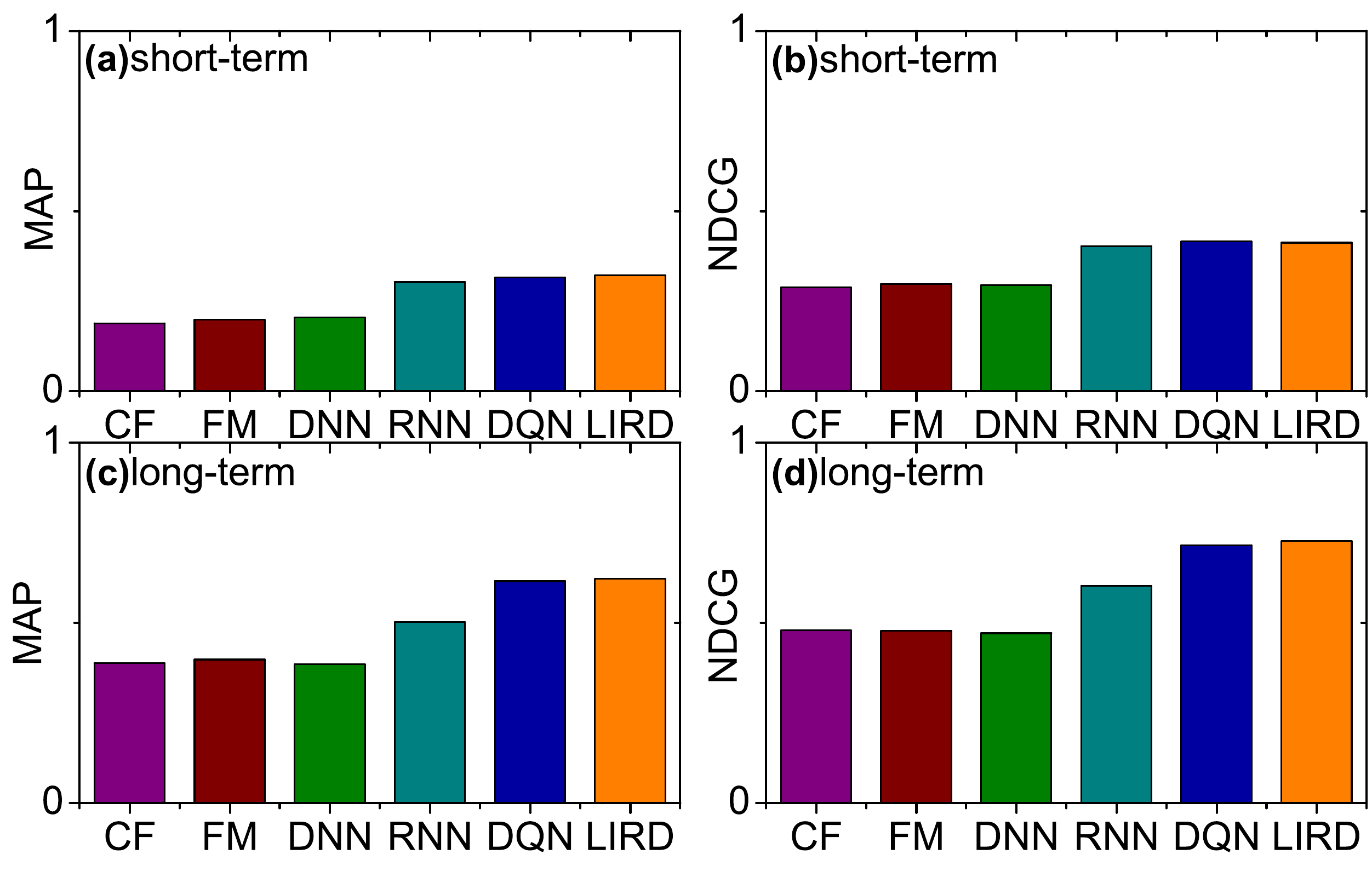}
	\caption{Overall performance comparison.}
	\label{fig:overall}
	\vspace{-3mm}
\end{figure}

As the testing stage is based on the simulator, we can artificially control the length of recommendation sessions to study the performance in short and long sessions. We define short sessions have less than 50 recommended items, while long sessions have more than 50 recommended items. The results are shown in Figure \ref{fig:overall}. We make following observations:
\begin{itemize}[leftmargin=*]
\item In both short and long sessions, CF, FM and DNN achieve worse performance than RNN, DQN and LIRN, since CF, FM and DNN ignore the temporal sequence of the users' browsing history, while RNN can capture the temporal sequence, DQN and LIRN are able to continuously update their strategies during the interactions.
\item In short recommendation sessions, RNN, DQN and LIRD achieve comparable performance. In other words, RNN models and reinforcement learning models like DQN and LIRD can both recommend proper items matching users' short-term interests.
\item In long recommendation sessions, DQN and LIRD outperforms RNN significantly, because RNN is designed to maximize the immediate reward for recommendations, while reinforcement learning models like DQN and LIRD are designed to achieve the trade-off between short-term and long-term rewards. This result suggests that introducing reinforcement learning can improve the performance of recommendations.
\item LIRD performs similar to DQN, but the training speed of LIRD is much faster than DQN, since DQN computes Q-value for all potential actions, while LIRD can reduce this redundant computation. This result indicates that LIRD is suitable for practical recommender systems with the enormous action space.
\end{itemize}

To sum up, we can draw the answer to the first question -- the proposed framework outperforms most representative baselines in terms of recommendation performance; while LIRD can be efficiently trained compared to DQN. 

\subsection{Performance of List-Wise Recommendations}

To validate the effectiveness of the list-wise recommendation strategy, we investigate how the proposed framework LIRD performs with the changes of the length of the recommendation list, i.e., $K$, in long-term sessions, while fixing other parameters. Note that $K =1$ is the item-wise recommendation.

\begin{figure}[t]
	\centering
	\includegraphics[width=81mm]{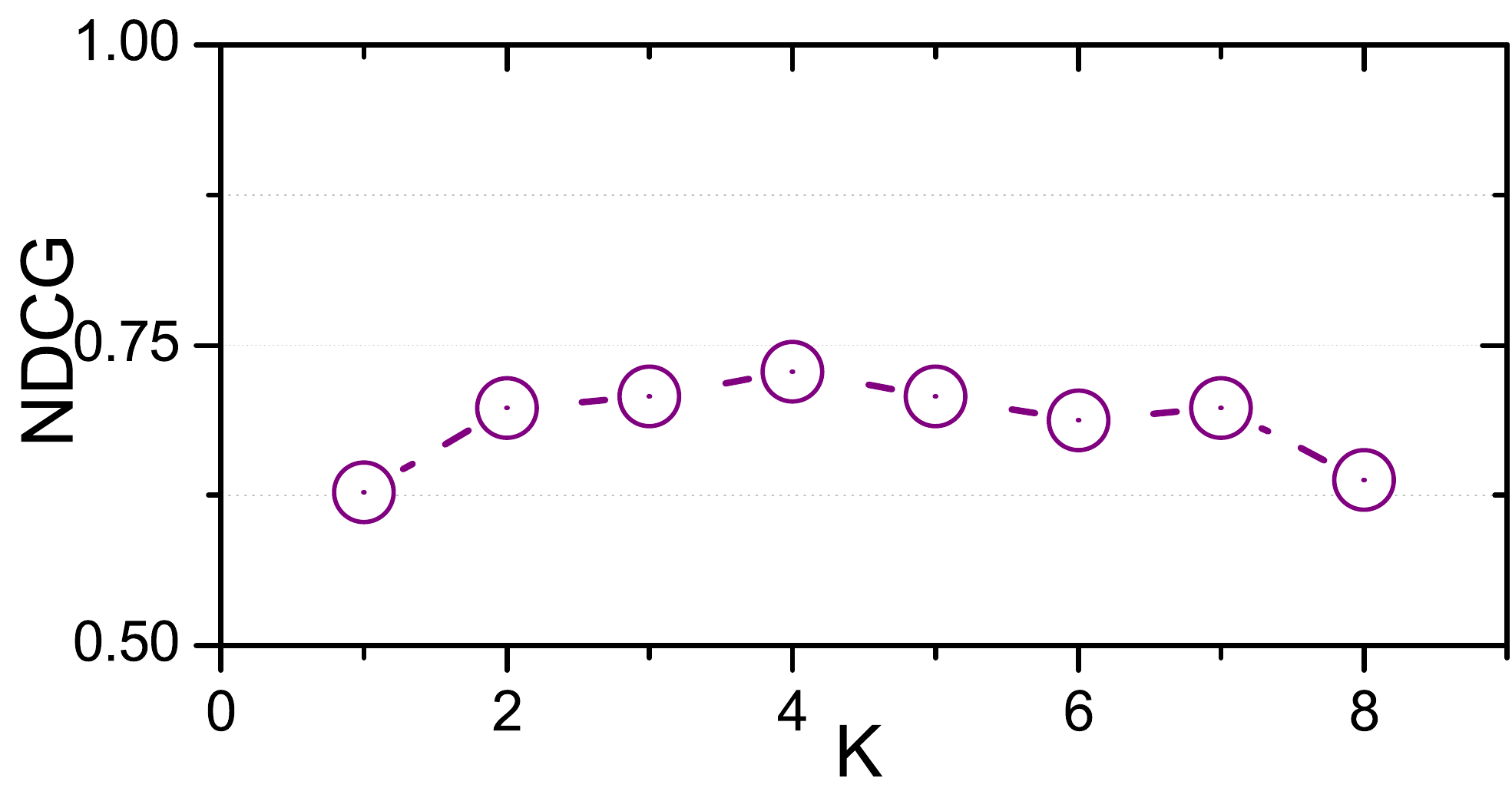}
	\caption{Performance with Recommendations Length $K$.}
	\label{fig:architecture}
\end{figure}

The results are shown in Figure \ref{fig:architecture}. It can be observed: 

\begin{itemize}[leftmargin=*]
\item In general, the recommendation performance first increases and then decreases with the increase of the length of the recommended list. 
\item The proposed framework achieves the best performance when $K = 4$. In other words, LIRD with a smaller $K$ could lose some correlations among the items in the same recommendation list; while the proposed framework with a larger $K$ will introduce noises. 
\end{itemize}

In summary, the list-wise recommendation strategy with appropriately selected $K$ can boost the recommendation performance, which answers the second question. 

\subsection{Performance of Simulator}
\label{sec:parametric}

The online simulator has one key parameter, i.e., $\alpha$, which controls the trade-off between state and action similarity in simulator, see Eq.(\ref{equ:probability1}). To study the impact of this parameter, we investigate how the proposed framework LIRD works with the changes of $\alpha$ in long-term sessions, while fixing other parameters.

The results are shown in Figure \ref{fig:overall}. We note that the proposed framework achieves the best performance when $\alpha = 0.2$. In other words, when we map current state-action pair $p_t (s_t, a_t)$ to a reward (the probability is based on the similarity between $p_t$ and historical historical state-action pair $m_i(s_i, a_i)$ in the memory), the action-similarity makes more contribution, while state-similarity also influences the reward mapping process.

\subsection{Discussion of Positional and Temporal order}
%
%
%

We build our LIRD framework under an assumption that in a page of recommended items, the user will browse items following positional order, i.e., user will observe the page of items from top to bottom. In this way the previous positional items can influence latter positional items, but not vice versa. However, sometimes users may add items to shopping cart, continue to browse latter positional items, and then make decision whether they order the items in shopping cart. Thus latter positional items could also influence previous items in reverse, i.e., positional order is not strictly equal to temporal order. We will leave it as one future investigation direction.

\begin{figure}[t]
	\centering
	\includegraphics[width=81mm]{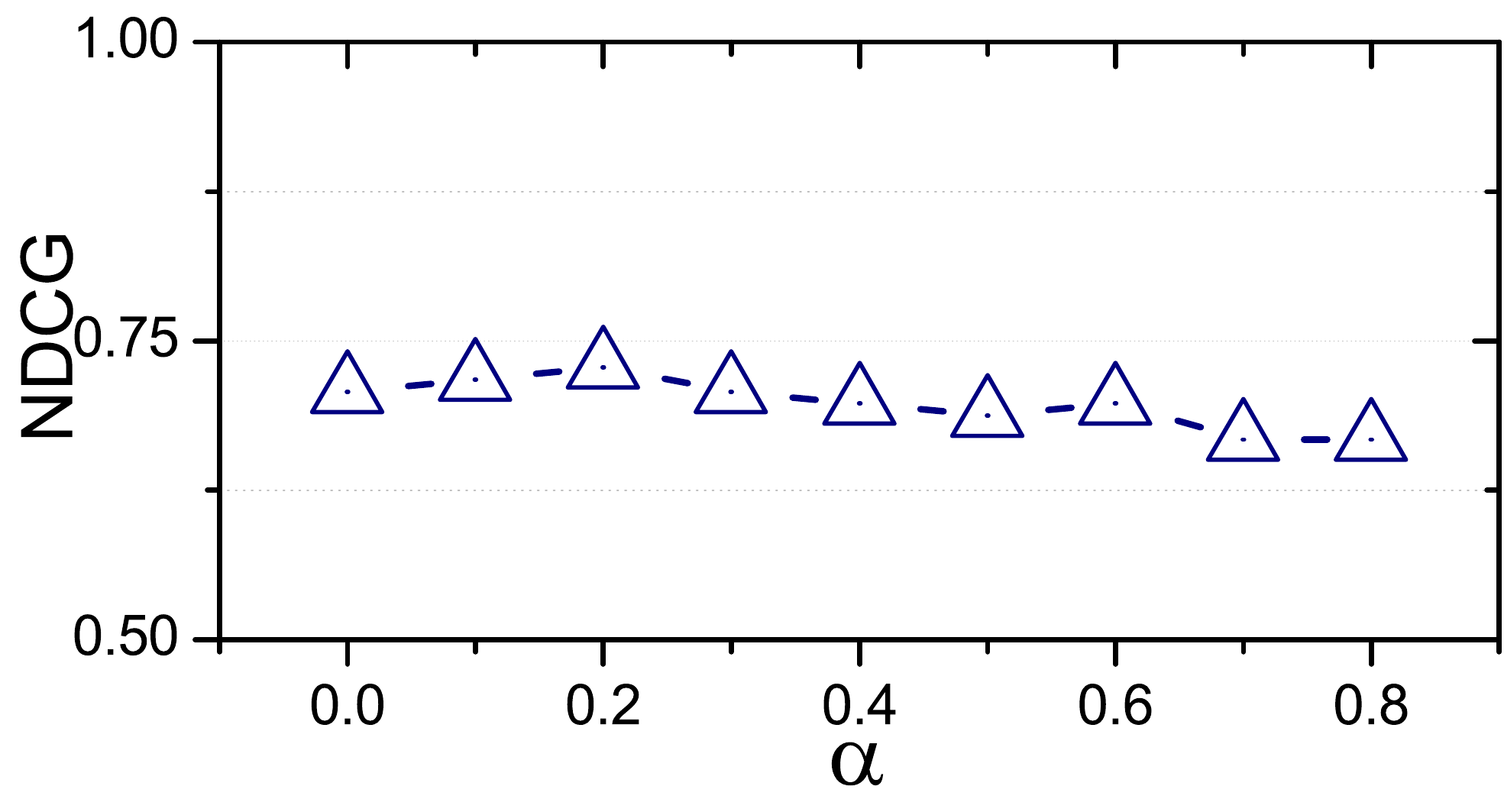}
	\caption{Parameter sensitiveness of $\alpha$}
	\label{fig:parameters}
\end{figure}
\section{Related Work}
\label{sec:related_work}

In this section, we briefly review works related to our study. In general, the related work can be mainly grouped into the following categories.

The first category related to this paper is traditional recommendation techniques. Recommender systems assist users by supplying a list of items that might interest users. Efforts have been made on offering meaningful recommendations to users. Collaborative filtering\cite{linden2003amazon} is the most successful and the most widely used technique, which is based on the hypothesis that people often get the best recommendations from someone with similar tastes to themselves\cite{breese1998empirical}. Another common approach is content-based filtering\cite{mooney2000content}, which tries to recommend items with similar properties to those that a user ordered in the past. Knowledge-based systems\cite{akerkar2010knowledge} recommend items based on specific domain knowledge about how certain item features meet users’ needs and preferences and how the item is useful for the user. Hybrid recommender systems are based on the combination of the above mentioned two or more types of techniques\cite{burke2002hybrid}. The other topic closely related to this category is deep learning based recommender system, which is able to effectively capture the non-linear and non-trivial user-item relationships, and enables the codification of more complex abstractions as data representations in the higher layers\cite{zhang2017deep}. For instance, Nguyen et al.\cite{nguyen2017personalized} proposed a personalized tag recommender system based on CNN. It utilizes constitutional and max-pooling layer to get visual features from patches of images. Wu et al.\cite{wu2016personal} designed a session-based recommendation model for real-world e-commerce website. It utilizes the basic RNN to predict what user will buy next based on the click histories. This method helps balance the tradeoff between computation costs and prediction accuracy.

The second category is about reinforcement learning for recommendations, which is different with the traditional item recommendations. In this paper, we consider the recommending procedure as sequential interactions between users and recommender agent; and leverage reinforcement learning to automatically learn the optimal recommendation strategies. Indeed, reinforcement learning have been widely examined in recommendation field. The MDP-Based CF model in Shani et al.\cite{shani2005mdp} can be viewed as approximating a partial observable MDP (POMDP) by using a finite rather than unbounded window of past history to define the current state. To reduce the high computational and representational complexity of POMDP, three strategies have been developed: value function approximation\cite{hauskrecht1997incremental}, policy based optimization \cite{ng2000pegasus,poupart2005vdcbpi}, and stochastic sampling \cite{kearns2002sparse}. Furthermore, Mahmood et al.\cite{mahmood2009improving} adopted the reinforcement learning technique to observe the responses of users in a conversational recommender, with the aim to maximize a numerical cumulative reward function modeling the benefit that users get from each recommendation session. Taghipour et al.\cite{taghipour2007usage,taghipour2008hybrid} modeled web page recommendation as a Q-Learning problem and learned to make recommendations from web usage data as the actions rather than discovering explicit patterns from the data. The system inherits the intrinsic characteristic of reinforcement learning which is in a constant learning process. Sunehag et al.\cite{sunehag2015deep} introduced agents that successfully address sequential decision problems with high-dimensional combinatorial slate-action spaces.  

\section{Conclusion}
\label{sec:conclusion}

In this paper, we propose a novel framework LIRD, which models the recommendation session as a Markov Decision Process and leverages Deep Reinforcement Learning to automatically learn the optimal recommendation strategies. Reinforcement learning based recommender systems have two advantages: (1) they can continuously update strategies during the interactions, and (2) they are able to learn a strategy that maximizes the long-term cumulative reward from users. Different from previous work, we propose a list- wise recommendation framework, which can be applied in scenarios with large and dynamic item space and can reduce redundant computation significantly. Note that we design an online user-agent interacting environment simulator, which is suitable for offline parameters pre-training and evaluation before applying a recommender system online. We evaluate our framework with extensive experiments based on data from a real e-commerce site. The results show that (1) our framework can improve the recommendation performance; and (2) list-wise strategy outperforms item-wise strategies. 

There are several interesting research directions. First, in addition to positional order of items we used in this work, we would like to investigate more orders like temporal order. Second, we would like to validate with more agent-user interaction patterns, e.g., adding items into shopping cart, and investigate how to model them mathematically for recommendations. Finally, the framework proposed in the work is quite general, and we would like to investigate more applications of the proposed framework, especially for those applications with both positive and negative(skip) signals. 

\bibliographystyle{ACM-Reference-Format}
\bibliography{6rl} 

\end{document}